# Developing Embodied Multisensory Dialogue Agents

Michał B. Paradowski[1]

**Abstract.** A few decades of work in the AI field have focused efforts on developing a new generation of systems which can acquire knowledge via interaction with the world. Yet, until very recently, most such attempts were underpinned by research which predominantly regarded linguistic phenomena as separated from the brain and body. This could lead one into believing that to emulate linguistic behaviour, it suffices to develop 'software' operating on abstract representations that will work on any computational machine. This picture is inaccurate for several reasons, which are elucidated in this paper and extend beyond sensorimotor and semantic resonance. Beginning with a review of research, I list several heterogeneous arguments against disembodied language, in an attempt to draw conclusions for developing embodied multisensory agents which communicate verbally and non-verbally with their environment.

Without taking into account both the architecture of the human brain, and embodiment, it is unrealistic to replicate accurately the processes which take place during language acquisition, comprehension, production, or during non-linguistic actions. While robots are far from isomorphic with humans, they could benefit from strengthened associative connections in the optimization of their processes and their reactivity and sensitivity to environmental stimuli, and in situated human-machine interaction. The concept of multisensory integration should be extended to cover linguistic input and the complementary information combined from temporally coincident sensory impressions.

**Keywords:** embodiment, sensorimotor resonance, semantic resonance, language, multisensory integration, robotics

## 1 INTRODUCTION

> *... His eyes only see*
> *His ears only hear ...*
> —Wisława Szymborska *No End of Fun* (1967)

In the 'traditional' view, going back to René Descartes, cognition has been seen as manipulation of symbolic, mental representations, with the brain conceived of as an input-output processor, a problem-solving device running abstract, generalised computational programs which enable us to process incoming data into a perception/interpretation of the outside world. This 'software', separate from the body, was equated with the mind, while the body was regarded as an output system attached to the cognitive processing system, with similar tasks achieved by applying the same underlying motor program to different effectors:

*magnam esse differentiam inter mentem & corpus, in eo quǹd corpus ex naturâ suâ sit semper divisibile, mens autem plane indivisibilis ... mentem a corpore omnino esse diversam.*[2]
 —Descartes (1641) *Meditationes de prima philosophia VI*;19

The information-processing approach or computer metaphor has become further entrenched over the latter half of the previous century due to the adoption of the digital computer as the platform to run the symbolic computations (Hoffmann *et al.*, n.d.).

However, this dualist perspective has been increasingly challenged, beginning with Edmund Husserl, Martin Heidegger, John Dewey, and Maurice Merleau-Ponty, and it is today widely acknowledged that perception and cognition are grounded in bodily experience. The brain is not the sole problem-solving resource we have at our disposal; the organiser/filtering machine is the *body-en-total*. Heuristics depend on our physiology; cognition is not only influenced and biased by states of the body, but crucial to it are also the rest of the body beyond the brain, as well as the environment.

Until very recently, most language research has, in a Cartesian manner, traditionally regarded linguistic phenomena as internal, mental, isolationist and amodal (that is, separate and independent from perception, action and emotion systems, and the body); a view endorsed in psychology (e.g. Geschwind 1970; Kintsch 1998), philosophy (e.g. Katz & Fodor 1963; Fodor 1983), and linguistics (e.g. early Chomsky – 1957, 1975; Nowak *et al.* 2002; Jackendoff 2002)[3]. For instance, Chomsky's most seminal theories were based on mathematical formalism and saw language as governed by a context-free grammar extended with transformational rules operating on (non-semantic) symbol strings and complemented by morphophonemic rules, with autonomous syntax at the core of the theory of language. The reason why his views for a long time did not go beyond such a perspective should not come as a surprise. His *Syntactic Structures*, which became a revolutionary and foundational work[4] in linguistics, grew out of a series of lecture notes for an audience of undergrad (mainly electrical engineering and maths) students at the MIT.[5] Also, Chomsky's ideas were born at the same time as the establishment of computer science as a distinct academic discipline, the beginnings of computational linguistics, and the founding of AI research, which all shared the dominant idea that thought can be described with formal logic.

---

[1] Inst. of Applied Linguistics, Univ. of Warsaw, ul. Browarna 8/10, PL-00-311 Warsaw, Poland. E-mail: `michal.paradowski@uw.edu.pl`.

[2] "There is a great difference between mind and body, inasmuch as body is by nature always divisible, and the mind is entirely indivisible. […] the mind or soul of man is entirely different from the body."

[3] A *votum separatum* in this domain is the field of biolinguistics, which hypothesizes a strong genetic (or neurobiological) endowment for language (UG) and determination of its structure (e.g. postulating selectional—i.e. evolutionary fitness—advantages), treating the language faculty on a par with other biological systems (see e.g. Meader & Muyskens 1950; Lenneberg 1967; Piatelli-Palmarini 1989; Hauser *et al.* 2002; Chomsky 2005; Di Sciullo & Boeckx 2011).

[4] Ranked #1 on the list of the one hundred most influential works in cognitive science from the 20th century, selected by the University of Minnesota Center for Cognitive Sciences: `http://www.cogsci.umn.edu/OLD/calendar/past_events/millennium/final.html`

[5] Before that, at the University of Pennsylvania, Chomsky studied logic and foundations of mathematics.

The generative school inspired several decades of linguistic thought, and even theories trying to modify or undermine its tenets were still relying on the underlying view of language as a system manipulating abstract symbols. This dualistic view could lead one into believing that in order to credibly emulate linguistic behaviour, it suffices to develop 'software' operating on (i.e. applying combinatorial rules such as Merge and Move to) abstract representations[6] that will work on any computational machine, and that its operations will be implementation-independent, functioning identically regardless of the physical hardware.

## 2 EMBODIED LANGUAGE IN HUMANS

*to turn print into exciting situations in their skulls*
—Kurt Vonnegut *Slaughterhouse-Five* (1969:205)

The dualistic approach just outlined above works to some extent in statistical machine translation, automatic text indexing and retrieval (think e.g. search engines), natural-language interfaces or dialogue systems, but if the system to be developed is to truly mimic human behaviour, the disembodied picture is not very accurate for several reasons. One may be doubtful about modularity and the existence of a specifically dedicated innate language acquisition device, but must still take into account the following phenomena and theoretical developments:

1. **lateralization** and **localization** of the language faculty in the brain. Linguistic capabilities have been shown to be limited to certain areas of the cerebrum, as evidenced primarily by various language disorders:[7]
   - receptive aphasia, commonly known as Wernicke's aphasia (Wernicke 1874): damage to the medial temporal lobe destroying local language regions and cutting them off from most of the occipital, temporal and parietal regions (cf. e.g. Price 2000; Bookheimer 2002; Damasio *et al.* 2004);
   - expressive aphasia (aka Broca's or agrammatic aphasia; Broca 1861);
   - abnormal language developed in individuals with the left hemisphere removed (Dennis & Whitaker 1976);[8]
   - Specific Language Impairment (SLI), which is unrelated to other developmental disorders, mental retardation, brain injury, or deafness (e.g. Joanisse & Seidenberg 1998; Bishop & Snowling 2004; Archibald & Gathercole 2006);
   - other cases of people with normal nonverbal abilities but impaired language, and 'normal' language but cognitive deficits (cf. the classic case studies of individuals with incommensurable linguistic and cognitive capacities: Genie (Curtiss 1981), Laura (Yamada 1990), Clive (Smith 1989), or Christopher (Smith *et al.* 1993)[9].

   While these deficits cannot straightforwardly be taken as proof of the *modularity* of language (*cf.* e.g. Calabrette *et al.* 2003; Fodor 2005), they do point to *localisation* of language processes;

2. embodiment of language in neuronal circuitry. FMRI studies have shown '**activation**' of certain brain areas involved in **language processing** (e.g. Osterhout 1997; Hagoort *et al.* 1999; Embick *et al.* 2000; Horwitz *et al.* 2003; Pulvermüller & Assadollahi 2007), with different levels of language processing identified in specific regions, e.g. loci of syntax mainly in left-perisylvian language regions, especially Broca's and Wernicke's areas, but also adjacent neocortical areas, the insula, and subcortical structures including basal ganglia (*cf.* e.g. Ullman 2001; Grodzinsky & Friederici 2006), or phonology in the superior temporal sulcus and anterior superior temporal cortex (*cf.* e.g. Diesch *et al.* 1996; Obleser *et al.* 2006; Uppenkamp *et al.* 2006);

3. genetic influence on language. While mutations of the Foxhead box protein 2 (**FOXP2** gene), deemed to cause a severe speech and language disorder (e.g. Lai *et al.* 2001; Vernes *et al.* 2008; Fisher & Scharff 2009), were initially taken as evidence for a 'language gene', it was later discovered that the protein impacts a wide range of phenotypic features all over the body (including facial motor control) and that the impairments of the family affected with the mutation went beyond language to other cognitive capacities. It is now more believed that it is networks of gene interactions rather than individual genes that have an influence on language (Knopka *et al.* 2009), but the neurobiological influence is there;

4. many Universal Grammar-based constraints now being reinterpreted as **learning** and **processing constraints**. That is, the difficulty in the acquisition of certain aspects of language are being accounted for by their complexity, the computational load under which the user/learner operates, his/her memory and attention limitations, or ease of access to representations (*cf.* e.g. Wakabayashi 2002; Van Hell & De Groot 1998; Wątorek 2008);
5. maturation and the **critical/sensitive period**[10] (but consider e.g. Marinova-Todd *et al.* 2000 for a contradictory view);
6. the Chomskyan **competence** vs. **performance** distinction (Chomsky 1965)[11], explaining mistakes in (originally native) language users' output (i.e., their actual deployment of the linguistic capacity) attributable to such psychosomatic states and factors affecting them as fatigue, tedium, intoxication, drugs, sudden changes of mind, haste, inattention, or external distractions;
7. interaction between (context-bound) language comprehension and production, and sensorimotor activation, manifested in both directions by:[12]
   - **motor resonance** observed in linguistic (Lakoff & Johnson 1980; Lakoff 1987), behavioural (primarily with priming[13] modulating motor performance; e.g. Tanenhaus *et al.* 1995; Gentilucci *et al.* 2000; Spivey *et al.* 2001; Glenberg & Kaschak 2002; Glover *et al.* 2004; Buccino *et al.* 2005; Boulenger *et al.* 2008; Nazir *et al.* 2008; Frak *et al.* 2010; for grammar *cf.* Madden & Zwaan 2003; Bergen & Wheeler 2010), neuroimaging and TMS studies[14] (e.g. Zatorre *et al.* 1992; Fadiga *et al.* 2002; Tettamanti *et al.* 2008; Fischer & Zwaan 2008; Kemmerer *et al.* 2008; Boulenger *et al.* 2009; Willems *et al.* 2010; for activation in visual areas *cf.* Martin *et al.* 1996; Pulvermüller & Hauk 2006; Simmons *et al.* 2007; in the olfactory cortex *cf.* González *et al.* 2006);
   - **semantic resonance** (brain language areas getting activated during sensorimotor action; Bonda *et al.* 1994; Pulvermüller *et al.* 2005; Rueschemeyer *et al.* 2010);[15]
   - verbalization of memory facilitated when assuming the original body position during recall (Dijkstra *et al.* 2007)[16], linguistic tasks expedited when accompanied by action (Rieser *et al.* 1994), and sensorimotor experiences intertwined with cognition in episodic memory (Pfeifer 2011);
   - faster comprehension of depictions of spatial associations than of descriptions of spatial *dis*sociations[17] (Glenberg *et al.* 1987); speedier recognition of words with 'body-object interaction' than of ones without (Siakaluk *et al.* 2008);
   - semantic interference and facilitation in the Stroop effect (longer RTs needed to name colour names written in incongruent ink hue; Jaensch 1929; Stroop 1935);
   - clinical studies indicating that processing of action concepts degrades if action- or vision-related brain areas are lesioned in motor neuron diseases (Damasio *et al.* 1996; Bak *et al.* 2001; Neininger & Pulvermüller 2003) and semantic dementia (Pulvermüller *et al.* 2010);
   - comprehension of action words deteriorating after loss of procedural knowledge (*cf.* Boulenger *et al.* 2008 on Parkinson's disease patients; also Bak *et al.* 2006);
8. parallel emergence of speech and gesture in infancy (Iverson & Thelen 1999);
9. co-speech gesture reducing cognitive load (Goldin-Meadow *et al.* 2001), and indications of a dual-task advantage for bimodal (signed-spoken) language production (i.e., production of code-blends, with elements of the signed and spoken languages appearing simultaneously; Kaufmann & Kaul 2012); or
10. Conceptual Blending theory (Fauconnier & Turner 2002) explaining language creativity as a semantic process operating on the output of perception and interaction with the world to create new structures.

---

[6] Understood as terminal symbols, which can—subsequently or concurrently—be equipped with referential, meaning-bearing properties.

[7] Theoretically, an injury disrupting the system's functioning may only show the involvement of the affected region, not that the whole functionality was due to that region. However, interestingly, not only spoken, but also sing language is left-lateralised (with use of classical language areas—e.g. Broca's (Horwitz *et al.* 2003)—in sentence processing and LH damage associated with lexical comprehension, with a difference in more posterior activation in areas responsible for processing vision and movement; Woll 2012). While signing patients with *RH* damage perform within the normal range on language tests, with the exception of tests of locative sentence comprehension, these problems appear to mean not linguistic malfunction *per se*, but an indirect consequence of more general cognitive deficits: in areas such as classifiers, spatial verbs, and grammar relying on space, sign language processing is reliant on visuospatial cognition (*ibid.*; Woll & Morgan 2012).

[8] Although one must be cautious about the conclusions since the cortical development in the subjects of the study was not normal in the first place (Chomsky 1980:264).

[9] In short, Christopher was able to acquire natural languages (with great aptitude, too, especially regarding morphology), but not ones violating the constraints of Universal Grammar. (The picture is more complex, but does not invalidate the basic claim.) But see e.g. Karmiloff-Smith (1998), Johnson *et al.* (1999), or Elsabbagh & Karmiloff-Smith (2006) for reports on Williams syndrome questioning evidence for a clear-cut dissociation of innate mechanisms for language. While the syndrome was originally postulated as characterised by preserved language in the presence of marked visual-spatial impairments, hence as evidence for modularity (*cf.* e.g. Bellugi *et al.* 1988, 1994), it was subsequently observed that actually language is not wholly intact (e.g. involving prepositional errors; Rubba & Klima 1991, Capirci *et al.* 1996, Volterra *et al.* 1996; Karmiloff-Smith *et al.* 2003; Woll 2012).

Thus, independently of theoretical persuasion, without taking into account both the architecture of the human brain, and

---

[10] The Critical Period Hypothesis (or its idea), proposed by Penfield and Roberts (1959), posits the existence of an ideal window of time during which genetically endowed language acquisition can—given adequate stimuli—take place spontaneously, relatively effortlessly, and characteristically meeting a high degree of success, after which acquiring a language naturally, automatically and with complete ultimate attainment becomes impossible. "The earlier the better" rule of thumb captures the negative correlation between the age of acquisition onset and subsequent asymptotic attainment. Most evidence to support the claim was supplied by Eric Lenneberg (1967) in his *Biological Foundations of Language*. While the existence of a critical period is widely accepted where first language acquisition is concerned, attempts to extend it to second language acquisition still arouse a good deal of contention (for instance, Lamendella (1977) suggested the term 'sensitive period' to emphasise the fact that acquisition may be more efficient during childhood, but not restricted to that period).

[11] The distinction can be considered on the example of any organic system: "Studies of the digestive system, for example, distinguish between its structural properties and what it is doing after you ate a sandwich" (Noam Chomsky, p.c., 8 Nov 2011), and can actually be traced back to the classic Aristotelian dichotomy between δύναμις (potentiality) and ἐνέργεια (actuality).

[12] This seems to be a reflection of a more general phenomenon where "there is no animal in which there is known to be a complete segregation of sensory processing" (Stein *et al.* 1996:497).

[13] E.g. in the form of mention of tool and action concepts.

[14] Somewhat importantly, motor resonance was not observed when the stimuli were used in idiomatic contexts (Rueschemeyer *et al.* 2010a) or metaphorical ones. Regarding the latter, Raposo *et al.* (2009) found activity in the pre- and motor cortex for literal-only usages of arm- and leg-related Vs, while Bergen *et al.* (2007) likewise demonstrated that visual imagery is triggered in sentence comprehension tasks (where general words of motion were employed) only where the utterances have literal spatial meaning. However, the picture is not completely clear-cut. This year, Lacey *et al.* (2012) showed that textural metaphors do activate parietal operculum regions important to the sense of touch. To explain this discrepancy, one could posit a qualitative difference between 'directly' embodied sensory experiences (e.g. texture or temperature) and more 'indirect' ones such as those grounded in visual perception. The former are more 'primary':
  i) sensed earliest – already in the womb, taction being the first sense that begins to develop before 8 weeks gestational age together with the emergence of the nervous system (Montagu 1978), before taste and smell (14 weeks g.a.), hearing (16 weeks g.a.; Shahidullah & Hepper 1992) or vision (week 18 onwards),
  ii) available in more 'primitive' organisms without vision or hearing,
  iii) perceptible during half-sleep, and
  iv) impacting our bodily functioning more strongly (the somatic reaction to extremely high or low temperatures, pressure or skin irritation is more likely to be stronger than e.g. to an unpleasant sight or sound).
This might account for the lack of activation in visual cortical areas.

[15] But see e.g. Bedny *et al.* (2008), Postle *et al.* (2008), or Kemmerer & Gonzalez-Castillo (2010) for opposing views.

[16] This conviction can also be found in 'folk wisdom'. For instance, in one episode of a Malaysian edutainment program for children which I was consulting on for a European broadcaster, a monkey was hanging upside down because that was the position in which she last saw her orange juice.

[17] I.e. texts describing an event in which the main character was spatially dissociated from a target object, e.g.:
```
John was preparing for a marathon in August. After
doing  a  few  warm-up  exercises,  he  took  off  his
sweatshirt and went jogging. (emph.added)
```

embodiment—the interaction of the language faculty with the sensory apparatus and motor system—it is unrealistic to replicate accurately the processes which take place during language acquisition, comprehension, or production, or during non-linguistic actions. Cognitive mechanisms are synergistically intertwined with affective and somatic components, and largely inseparable (Ziemke 2011).

## 3 THE COROLLARIES FOR ROBOTICS

*… it is the movement which is primary, and the sensation which is secondary, the movement of the body, head, and eye muscles determine the quality of what is experienced.*
*In other words, the real beginning is with the act of seeing; it is looking, and not a sensation of light.*
—John Dewey (1896:358*f.*)

Since the official launch of AI as a new research discipline at the seminal Dartmouth conference in 1956, much of work in the field has been driven by the 'Physical Symbol Hypothesis' (Newell & Simon 1976): trying to construct systems that would possess or build internal, symbolic representations of objects and relations in the outside *world*—in other words, a "world model"—which usually had little to do with their hardware, sensorimotor *experience*, or current context[18], but were instead characterised by precisely defined states and finite lists of acceptable commands (Wang 2009:2*f.*). Under such a functionalist approach, the body is merely a platform on which cognitive operations are running. In some areas, such closed systems were able to achieve spectacular feats, for instance in defeating world chess champions.

Chess, however, is a formal game, set in a virtual world with discrete states, positions, and licit moves, a game involving complete information, and a static one: no move means no change, and the inventory of legitimate operations remains constant (Pfeifer & Scheier 1999:58*ff.*). This is quite unlike what usually happens in the real world. Hence, the last two and a half decades have witnessed recurrent appeals for situated, embodied autonomous systems actively and directly interacting with the world around (*cf. op. cit.*; Brooks 1991; Varela *et al.* 1991) and constructing knowledge via this dynamic enactment (the active learning being qualitatively different from statistical machine learning; *cf.* e.g Froese 2009; Vernon 2010). Evidently robots, even anthropomorphic ones, are far from isomorphic with humans in terms of both the 'brain' and the rest of the body, including the input and output devices (sensors and actuators). Also, as one reviewer rightly remarks, in the language technology field priority is not necessarily to make a machine as humanlike as possible, with the same architecture; rather, it is to make the machine so that it does things on a level comparable to humans (or, I would add, surpassing that) – in other words, to achieve similar—or better—functionality in terms of mode, scope, or scale. Or, going completely beyond the anthropocentric GOFAI perspective (Haugeland 1985; *cf.* Wang 2008), since passing the Turing Test is not a *sine qua non* of being intelligent, as acknowledged by the test's designer himself (Turing 1950). This, however, means that robust artificial cognitive agents can bypass the human limitations[19] inherent in most of the above points (just as they could overcome some contingencies resulting from the material properties of the human brain and bodily features such as synaptic speed and efficiency, the physical characteristics of the vocal tract, the auditory perception system, or muscular flexibility[20]). Nevertheless, they could still benefit from strengthened associative connections owing to the motor and semantic resonance in both the optimization of their processes, and reactivity and sensitivity to environmental stimuli, across a range of tasks:

(i) in grounded language understanding (*cf.* e.g. Glenberg & Kaschak 2002; Feldman & Narayanan 2004; Gallese & Lakoff 2005; Sato *et al.* 2008), where structuring the environment acts as scaffolding[21] and all inputs contribute to evidential support,

(ii) in automated articulation-based speech recognition (utilising motor information, i.e. combining spoken input with visual data—e.g. the shape of the speakers lips—and maybe even data such as strength of the incoming airstream),

(iii) while learning about context-dependent phenomena in the surrounding world (e.g. action sequences and argument structure in construction grammar; *cf.* Dominey 2007; since embodiment plays a constitutive role in the process of cognition; Vernon 2010), or in the process of language acquisition in general (because language—at least in the initial stages—is acquired by situated embodied direct engagement with the world, and not just passive perception, e.g. watching television; *cf.* e.g. Steels 2009),

(iv) to help with storage and retrieval due to the benefits of episodic memory,

(v) to support action prediction, planning and anticipation (Koelewijn *et al.* 2008; Stapel *et al.* 2010; van Elk *et al.* 2010)[22], including prediction of the next sensory feedback,

(vi) to support action execution (with linguistic input making the actor better aware of the affordances, i.e. physically feasible action possibilities), and

(vii) to reinforce feedback in 'soft robotics' and morphological computation, where there is no clear separation between the controller (or orchestrator) and the hardware (morphology), and the tasks are distributed between the brain, body, and environment (*cf.* e.g. Paul

---

[18] The fact that the appropriate relations to some outside world could be established by the system's designer or end-user becomes unhelpful the moment we want to deal with an autonomous agent, with the human interpreter removed from the loop, as emphasised by Steven Harnad in his seminal (1990) paper (*cf.* also Pfeifer & Scheier 1999:69*f.*).

[19] The limitations need not in themselves necessarily be a bad thing; to the contrary, they may serve a useful role in limiting the search space and focusing attention on the most vital stimuli. The restrictions imposed on the vocal apparatus in turn mean that speech is segmented and decelerated enough to facilitate comprehension. The relative absence of such constraints on computers may be the exact reason why the latter have problems tackling tasks where humans perform with ease (Tom Froese, p.c., 9 Mar 2012).

[20] Just as robots can have an advantage when equipped with e.g. infrared, or ultrasonic sensors.

[21] Sensorimotor dynamics plays a crucial part in toddlers' learning to categorise objects: it is only when the infant brings the object in front of their eyes and focuses on it that s/he learns to associate it with its name (Smith 2010).

[22] Though originally grounded in sensorimotor experience, mental imagery, or simulation of interaction with the world, may subsequently become environmentally decoupled, as in forward models (Clark & Grush 1999).

2004; Pfeifer 2011; which also has the aim of off-loading computation; Di Paolo 2009);[23]

(viii) in cognitive developmental robotics, aiming at understanding human cognitive developmental processes by synthetic or constructive approaches (Asada *et al.* 2009, Asada 2011, Ishiguro *et al.* 2011);

(ix) in common grounding and alignment, which are crucial for fruitful situated human-machine interaction, and which are another area where sensory experience must be coordinated with linguistic interaction.

Principally, if our goal were to create machines which do things on a *comparable level* to—or surpassing—humans, we could do away with attempts at embodying them in human-inspired ways (Taivo Lints, p.c., 31 May 2012) – they could function perfectly well with totally nonhuman kinds of embodiment (different 'bodies', different sensors and effectors, different internal architectures... or even with embodiment in a virtual world; Bringsjord *et al.* 2008; Goertzel *et al.* 2008). Given the role played by the morphology of the sensory apparatus and the architecture of the sensorimotor loop in shaping and structuring the information that reaches the controller, and thereby in concept formation, it would anyway be difficult for a machine to form the same concepts, categories and behaviours as us without having comparable morphology (as remarked e.g. by Barsalou 1999 or Lakoff & Johnson 1998). However, if our goal is to have machines 'thinking' and behaving in a way *compatible* with ours—which is a highly practical and desirable goal—then it is of high importance for them to develop, learn and function in a similar "experience space" (Taivo Lints, p.c.; *cf.* also Wang 2009:5).

The requirement that the behaviour, perception and conceptual apparatus of artificial intelligent agents be grounded in their experience of their *own* interaction with the outside world at once means that their concepts and categories need not necessarily rely on the same minimal constituents and grammatical categories as have been externally identified and defined in linguistics. Instead, the gradually emergent categories are more likely to be intrinsically meaningful behaviours and affordances (see also Kuniyoshi *et al.* 2004), action-oriented rather than orbocentric (Hoffmann & Pfeifer 2011). For instance, to a robot who has never kicked or observed anyone kick anything but footballs, the minimal unit of meaning may be <kick a ball> rather than <kick> alone (although this does not rule out the possibility of extrapolation and abstraction should a relevant opportunity arise).[24] Similarly, irrespective of whether the input is expressed using [$_{NP}$ kicking a ball] or [$_{VP}$ kick a ball], it should activate the same action schema.

## 4 TOWARDS A BROADER DEFINITION OF MULTISENSORY INTEGRATION

[23] The idea of morphological computation in animals can be well illustrated on the example of cockroaches skilfully climbing over obstacles that exceed their body height, using relatively few neurons, off-loading most tasks to morphology (by reconfiguring the mesothoracic shoulder joint), exploiting mechanical change and feedback, and capitalising on the stability of the local feedback circuits; *cf.* Watson *et al.* 2002; Pfeifer *et al.* 2007; Pfeifer & Gomez 2009).

[24] See for instance the POETICON++ project (Robots need Language: A computational mechanism for generalisation & generation of new behaviours in robots; http://www.poeticon.eu/).

In order to form a meaningful experience and construct coherent, reliable and robust representations of the surrounding world, the human brain combines prior knowledge with sensory input arriving from various modalities and integrates these at multiple levels of the neuraxis. This serves to maximize the efficiency of everyday performance and learning, enhancing the salience of the events, helping increase the detection and identification of the external stimuli, disambiguate them, compensate for incomplete information, and shorten reaction times. In view of the inseparability of language and the body, the concept of multisensory integration—whether in natural or artificial cognitive agents—should be extended and cover both the linguistic input and the complementary information that the brain combines from temporally coincident sensory impressions. This does not mean that we should 'dumb down' the statistical processes where they operate successfully; instead, where the input stream in one channel is too noisy, turning on auxiliary channels[25] and interacting with the environment in an active manner may generate ancillary data and help e.g. disambiguate the signal and take the right decision (see also Pfeifer & Scheier 1997; Beer 2003).[26] An added benefit would then be significantly reduced programming costs.

## CONCLUSIONS

> *A living organism* enacts *the world it lives in; its effective embodied action in the world actually constitutes its perception and thereby grounds its cognition.*
> —Stewart, Gapenne & Di Paolo (2010:vii)

I have started out with a brief depiction of the dualistic Cartesian approach that has characterised much of twentieth-century thought, including that underlying most of traditional AI. While adherence to such an outlook has in many domains led to very spectacular achievements, there are limits which purely symbolic systems cannot overcome. While the subject of the mind-body relationship is by no means new, the link, still very often ignored by cognitive science communities (logic, linguistics, computer science) may be the key element for bypassing the present limitations of AI systems.

Language, too, has for a long time been treated across scientific domains as an abstract system operating largely independently from the body (articulatory-perceptual organs notwithstanding). I have presented an inventory of heterogeneous evidence against such a view, addressing instead the issue of the link between language and body. While many of the embodied language phenomena specific to humans have little direct translation to machines, there are others that can profitably be exploited and inspire the development of robust artificial autonomous agents that rely on semantics grounded in their past experience (both linguistic and non-verbal) as well as possible related operations on the concepts concerned. Agents which are adaptive to feedback and can, despite insufficient knowledge, time pressure and storage space constraints safely and successfully navigate, learn, and communicate in the complex and dynamic ecological niche they share with human actors.

[25] These channels need not all be active at all times, especially when it might burden the cognitive load in non-essential tasks, when conflicting inputs can bring the machine to a halt, or when the benefits—e.g. in terms of speed—would be negligible (Richard Littauer, p.c., 26 May 2012). The system's available resources should be dynamically allocated to different tasks in such a way as to achieve the highest overall efficiency.

[26] One consequence for humans may be that the role of kinaesthetic modality, traditionally largely believed to dominate in children, but be negligible in adults (*cf.* e.g. Barbe & Milone 1981; Felder & Spurlin 2005), should be reassessed, as the effectiveness may be demonstrated of 'learning-by-doing' and task-based approaches to language learning and teaching where the students have to use their bodies (e.g. when acquiring novel lexis via common cookery classes).

## ACKNOWLEDGMENTS

The author wishes to thank Noam Chomsky, Anna Esposito, Taivo Lints, Richard Littauer, Gary Lupyan, Vincent C. Müller, Katerina Pastra, Michael Pleyer, Yulia Sandamirskaya, Luc Steels, the Evolang IX reviewers, and the anonymous referees for this Symposium for invaluable commentary, discussion and bibliographical references. Naturally, willingness to comment does not imply endorsement; all the usual disclaimers apply. An earlier version of this paper was presented at the 4th EUCogII Members' Conference "Embodiment – Fad or Future?", Anatolia College/American College of Thessaloniki, 11-12 Apr. 2011.## REFERENCES